\pdfoutput=1
\documentclass[11pt]{article}
\usepackage[]{naacl2021}
\usepackage{times}
\usepackage{latexsym}

\usepackage{url}
\usepackage[T1]{fontenc}
\usepackage[utf8]{inputenc}
\usepackage{microtype}

\usepackage{multirow}
\usepackage{sidecap}
\usepackage{amsthm}
\usepackage{amsmath}
\usepackage{amssymb}
\usepackage{amsfonts}
\usepackage{verbatim} 
\usepackage{url}
\usepackage{pifont}
\usepackage{eqparbox}
\usepackage{arydshln} 
\usepackage{bigstrut}
\usepackage[linewidth=1pt]{mdframed}
\usepackage{framed}
\usepackage{float}
\usepackage{adjustbox}
\usepackage{booktabs}
\usepackage{physics}
\usepackage{url}
\usepackage[normalem]{ulem}
\usepackage{tabu}
\usepackage{subcaption}

\usepackage{comment}
\usepackage[colorinlistoftodos]{todonotes}

\definecolor{airforceblue}{rgb}{0.36, 0.54, 0.66}
\definecolor{implicit-color}{rgb}{0.56, 0.74, 0.56}
\definecolor{explicit-color}{rgb}{0.88, 0.66, 0.37}

\newcommand{\keywordCode}[1]{{\small \texttt{#1}}}

\usepackage{microtype}

\newcommand{\datasetname}{\textsc{Tracie}}
\newcommand{\modelpattern}{\textsc{PtnTime}}
\newcommand{\modelsymbolic}{\textsc{SymTime}}
\newcommand{\modelzeroshot}{\textsc{SymTime-ZeroShot}}
\definecolor{OliveGreen}{rgb}{0,0.6,0}
\usepackage{soul}
\definecolor{Gray}{gray}{0.9}
\definecolor{almond}{rgb}{0.94, 0.87, 0.8}
\definecolor{applegreen}{rgb}{0.4, 0.71, 0.0}
\DeclareRobustCommand{\hlgray}[1]{{\sethlcolor{Gray}\hl{#1}}}
\DeclareRobustCommand{\hlcyan}[1]{{\sethlcolor{applegreen}\hl{#1}}}

\title{
\vspace*{-0.5in}
{{\small \hfill NAACL'21}\\
\vspace*{.25in}}
Temporal Reasoning on Implicit Events from Distant Supervision
}
  
\author{Ben Zhou$^{1,2}$ ~ Kyle Richardson$^1$ ~ Qiang Ning$^3$ ~ Tushar Khot$^1$ ~ Ashish Sabharwal$^1$ ~ Dan Roth$^2$ \\
\thanks{~~Most of the work was done when the third author was employed at the Allen Institute for AI and the first author was an intern there.}  
{ \normalsize  $^1$Allen Institute for AI \;\;\;  $^2$University of Pennsylvania \;\;\; $^3$Amazon } \\
{\tt \footnotesize \{kyler,tushark,ashishs\}@allenai.org \; \{xyzhou,danroth\}@cis.upenn.edu \; qning@amazon.com}
}

\date{}

\begin{document} 
\maketitle

\begin{abstract}

We propose \datasetname{}, a novel temporal reasoning dataset that evaluates the degree to which systems understand {\em implicit} events---events that are not mentioned explicitly in natural language text but can be inferred from it. This introduces a new challenge in temporal reasoning research, where prior work has focused on explicitly mentioned events. Human readers can infer implicit events via commonsense reasoning, resulting in a more comprehensive understanding of the situation and, consequently, better reasoning about time. We find, however, that state-of-the-art models struggle when predicting temporal relationships between implicit and explicit events. To address this, we propose a neuro-symbolic temporal reasoning model, \modelsymbolic{}, which exploits distant supervision signals from large-scale text and uses temporal rules to combine start times and durations to infer end times. \modelsymbolic{} outperforms strong baseline systems on \datasetname{} by 5\%, and by 11\% in a zero prior knowledge training setting. Our approach also generalizes to other temporal reasoning tasks, as evidenced by a gain of 1\%-9\% on MATRES, an explicit event benchmark.


\end{abstract}

\section{Introduction}
\label{sec:introduction}

\begin{figure}[t]
    \centering
    \includegraphics[scale=0.49]{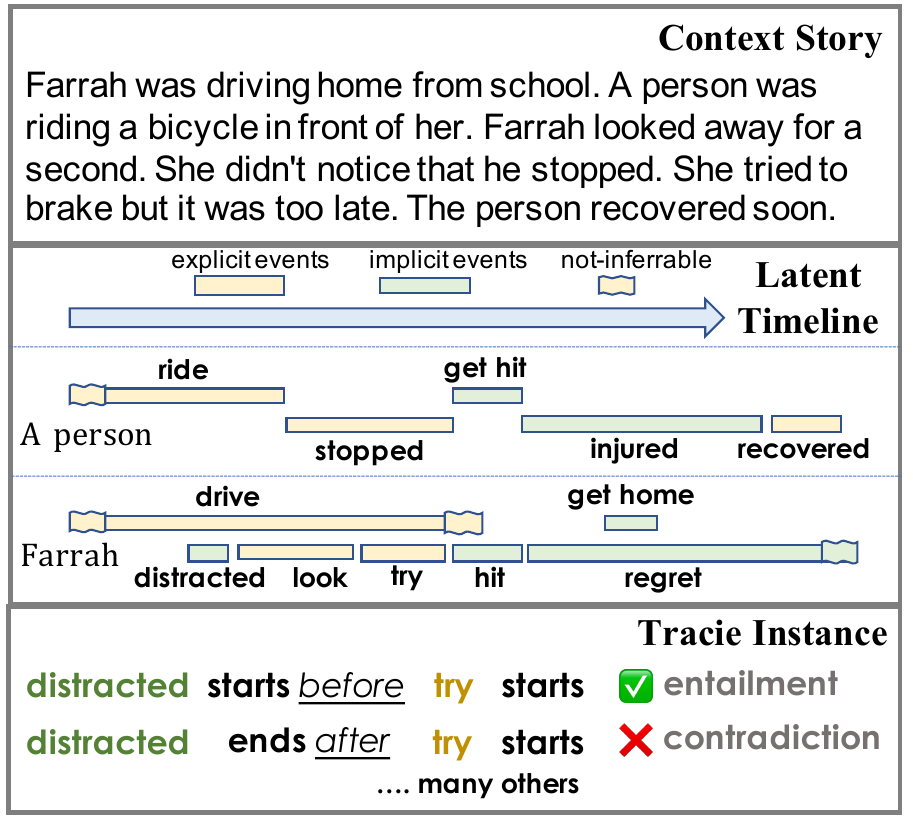}
    \caption{
        A story, its latent timeline, and example \datasetname{} instances from it. For simplicity, 
        events are shortened to single verbs and the timeline is exaggerated.
    }
    \label{fig:timeline-example}
\end{figure}

Understanding temporal relations between events in narrative text is a crucial part of text understanding.
When reading a story, a human can construct a latent timeline about events' start and end times, similar to the one shown in Fig.~\ref{fig:timeline-example} about an automobile accident. This timeline not only contains the placements of explicitly mentioned events (e.g., \emph{ride a bicycle}), but also accounts for implicit events (e.g., Farrah was \textit{distracted} so she looked away). Such a latent timeline explains the dynamics between events; for example, the possible chain of events between \textit{ride} and \textit{recovered} in this context contains \textit{get hit} and \textit{injured}. 
The ability to construct such a timeline is essential for understanding the causal dynamics of a situation. Without it, NLP systems cannot truly understand situations and reliably solve tasks such as temporal question-answering, causal inference, and scheduling assistance.

To better evaluate this ability, we introduce a new dataset called \datasetname{} (\emph{TempoRAl Closure InfErence}) that focuses on temporal relations on implicit events in short stories. Our dataset contains high-quality annotations of both start and end time queries that test a system’s understanding of the full temporal closure (i.e., both start and
end time) of events.
As a task that requires considerable commonsense knowledge, we follow \citet{zhou-etal-2020-temporal} in minimizing the size of the training set, therefore making \datasetname{} mainly an evaluation set.
The final \datasetname{} dataset contains a total of 5.4k human-curated instances, provided in a (multi-premise) textual entailment (TE) format, as illustrated at the bottom of Fig~\ref{fig:timeline-example}. A Pre-trained language model such as T5-Large \cite{raffel2020exploring} fine-tuned on our new dataset achieves a modest binary prediction accuracy of $67.9$\%.\footnote{The same model achieves 77.4\% on MATRES \cite{NingWuRo18} with a similar amount of training instances.
All \datasetname{} numbers reported in this section are from Table~\ref{tab:performance-iid-balanced}.
} 
Consistent with other studies on temporal reasoning \cite{zhou-etal-2020-temporal}, these results reveal serious limitations in existing pre-trained language models.


To build models better capable of understanding time with minimal direct training data, 
we propose a novel distant supervision technique that improves generalization by extracting temporal patterns in large-scale free text as part of an additional pre-training step. 
In contrast to other attempts at extracting temporal data through patterns at a sentence level \cite{gusev2011using, zhou-etal-2020-temporal}, we extract over large windows of text such as paragraphs. This allows for capturing global information related to multiple events and extracting signals that do not appear in small-window local contexts. The resulting model, \modelpattern{} (PatternTime), achieves a $76.6$\% accuracy on \datasetname{}, a $9\%$ gain over using standard T5-Large.
We also show the applicability of \modelpattern{} on a standard temporal reasoning benchmark involving only explicit events, MATRES \cite{NingWuRo18}, with a $9$ point gain in a low-resource setting.

We achieve further improvements by coupling \modelpattern{} with a duration model from \citet{zhou-etal-2020-temporal} to create a neural-symbolic reasoning model called \emph{\modelsymbolic{}}. The key idea in \modelsymbolic{} is to \emph{decompose} the computation of temporal relations to the predictions of relative distances between start times and those of durations.
For example, in Fig~\ref{fig:timeline-example}, we can decide that \emph{distracted} likely ends before \emph{try} starts because the duration of \emph{distracted} is likely to be shorter than the distance between the two start times.
This allows for better prediction on the end time, which rarely appears in the natural text and has been previously shown to be difficult to annotate \cite{NingWuRo18}. Such a symbolic computation involves a logical combination of the individual models in a way that formalizes part of the Allen interval algebra \cite{allen1983maintaining}. This model, which supports a wider range of temporal computation and can be used with and without task-specific supervision, achieves a final accuracy of $78.9$\%
on \datasetname{}'s binary classification metric. We also show that \modelsymbolic{} is more robust to different distributions of the training data, demonstrating the benefits of using a temporal model with a transparent reasoning process.


In summary, we make the following 3 contributions: (1) a temporal relation dataset \datasetname{} focusing on implicit events (\S\ref{sec:tracie}); (2) a distant supervision process for temporal understanding of implicit events (\S\ref{sec:pattern-based-pretraining}); and (3) a reasoning model that makes end-time comparisons using predictions of start-time distances and durations (\S\ref{sec:symtime}). Finally, we demonstrate the effectiveness of our models on \datasetname{}, as well as the applicability of our approach to an existing temporal benchmark (\S\ref{sec:experiments}).




\section{Related Work}


Temporal reasoning has received much attention in the NLP community, and to date, there are many datasets that focus on temporal ordering \cite{Pustejovsky2003TimeMLRS, Bethard2007TimelinesFT, Cassidy2014AnAF, reimers2016temporal, ogorman-etal-2016-richer, NingWuRo18, Ning2020TORQUEAR}, and other temporal knowledge \cite{Pan2006ExtendingTW, zhou2019going}. We focus here on modeling implicit events, which has received relatively little attention. Multiple systems have been proposed as part of research into temporal ordering \cite{Do2012JointIF, Moens2017StructuredLF, Leeuwenberg2018TemporalIE, Meng2018ContextAwareNM, Ning2018CogCompTimeAT, Han2019JointEA}, duration prediction \cite{vashishtha2019fine} and other tasks. Our decision to use a textual entailment style follows recent work on natural language inference \cite{williams2017broad,nie2019adversarial,bhagavatula2019abductive}, which tends to not focus on time (for recent work on temporal NLI, see \citet{vashishtha2020temporal}).
Many have used distant supervision for temporal reasoning \cite{gusev2011using, Ning2018ImprovingTR, zhou-etal-2020-temporal}. Comparatively, our work captures longer-range dependencies in narrative text (for related ideas, see \citet{ammanabrolu2020automated}). 

We are inspired by structural predictions and constraints that combat the sparsity of temporal knowledge \cite{Ning2017ASL, Do2012JointIF}, as well as neural module networks \cite{Andreas2016NeuralMN,gupta2019neural} and other decomposition-based approaches \cite{talmor2018web, KKSR18, li2019logic, wolfson2020break, khot2020text}. 
In particular, we build neural-symbolic transformer
models that operationalize some of the classical
interval-based computations used in earlier work
on temporal reasoning \cite{allen1983maintaining,gerevini1995efficient} (for related ideas, compare with \citet{Leeuwenberg2018TemporalIE,vashishtha2019fine}).




\begin{figure*}
    \centering
    {\footnotesize
            \begin{tabular}{| p{7.8cm} | p{4.1cm} | l |}
                 \hline 
                 \multicolumn{1}{|c}{\textbf{Context Story} (Premise)} & \multicolumn{1}{c}{\textbf{Hypothesis}} & \multicolumn{1}{c|}{\textbf{Inference Label}} \\ \hline
                 \emph{Tom needed to get braces. He was afraid of them. The dentist assured him everything would be fine. Tom had them on for a while. Once \hlgray{removed} he felt it was worth it.} & Tom avoids foods he can't eat with braces \hl{\texttt{starts}} \hlcyan{\texttt{before}} \hlgray{the braces are removed}. & \texttt{entailment} \\ \hline 
                 \emph{We were all \hlgray{watching Spongebob as a family}. It is a kid's show but all really enjoyed it. This one episode was especially funny for the adults. It has humor in it that is funny for kids and adults. It is something we can all watch...} & The adults laughed at the jokes \hl{\texttt{ends}} \hlcyan{\texttt{before}} \hlgray{we watch Spongebob as a family} & \texttt{contradiction} \\ \hline 
                 \emph{I was throwing the baseball with my son. He threw one past me that landed in the lake. \hlgray{I reached in to get the ball}. I lost my balance and fell in. I got the ball and a bath all in one shot!} & The ball was in the boys hand \hl{\texttt{starts}} \hlcyan{\texttt{after}} \hlgray{he reached for the ball} & \texttt{contradiction} \\ \hline
            \end{tabular}}
    \caption{Example \datasetname{} instances. The 
    \hl{\textbf{comparator}}  $l \in $\{\texttt{starts},\texttt{ends}\} and \hlcyan{\textbf{relation}} $r \in$\{\texttt{before},\texttt{after}\} in each hypothesis are highlighted, in addition to the corresponding \hlgray{explicit event} from the story.} 
    \label{fig:tracie_instances}
\end{figure*}

This work is broadly related to works on causal dynamics \cite{Pearl2009CausalII}. The nature of combined temporal and causal focuses is also related to procedural text modeling \cite{Tandon2018ReasoningAA,tandon2020dataset}.

\section{The TRACIE Dataset}
\label{sec:tracie}

In this section, we introduce the \datasetname{} dataset.\footnote{We release \datasetname{} and its leaderboard at \url{https://leaderboard.allenai.org/tracie}}

\subsection{Task Overview and Dataset Construction}
\label{subsec:tracie-overview}

The goal of \datasetname{} is to test a system's ability to compare start and end times of non-extractive implicit event phrases instead of extractive triggers from the context. Such tests in \datasetname{} take the form of multi-premise textual entailment (TE) \cite{lai2017natural}. Each \datasetname{} instance contains 1) a \textbf{context story} (or premise) consisting of a sequence of \emph{explicit} narrative events; 2) an \textbf{implicit event} in the form of a natural language phrase that is unmentioned but has some role in the story; 3) a \textbf{comparator} of either \texttt{\{starts,ends\}}; 4) an \textbf{explicit event} also in the form of a phrase, and 5) a \textbf{temporal relation} of either \texttt{\{before,after\}} that marks the relationship in the dimension defined by the \textit{comparator} between the \textit{implicit-event} and the \textit{explicit-event}. With these 4 components, we are able to generate TE-style instances, using the context story as the premise and temporal queries about pair-wise relations between implicit and explicit events as hypotheses.
For example, in the first positive instance shown in Fig.~\ref{fig:timeline-example}, ``distracted'' is the \textit{implicit-event}, ``starts'' is the \textit{comparator}, ``try'' is \textit{explicit-event} and ``before'' is the \textit{temporal-relation}. They form a positive hypothesis ``distracted starts before try.''\footnote{All event phrases are shortened to triggers here for simplicity. See Fig.~\ref{fig:tracie_instances} for actual phrases.} We flip the \textit{temporal-relation} (i.e., ``before'' to ``after'' and vice versa) to create negative (contradiction) instances, as shown in the second example instance in Fig.~\ref{fig:timeline-example}.

\begin{figure}[t]
    \centering
    \includegraphics[scale=0.34]{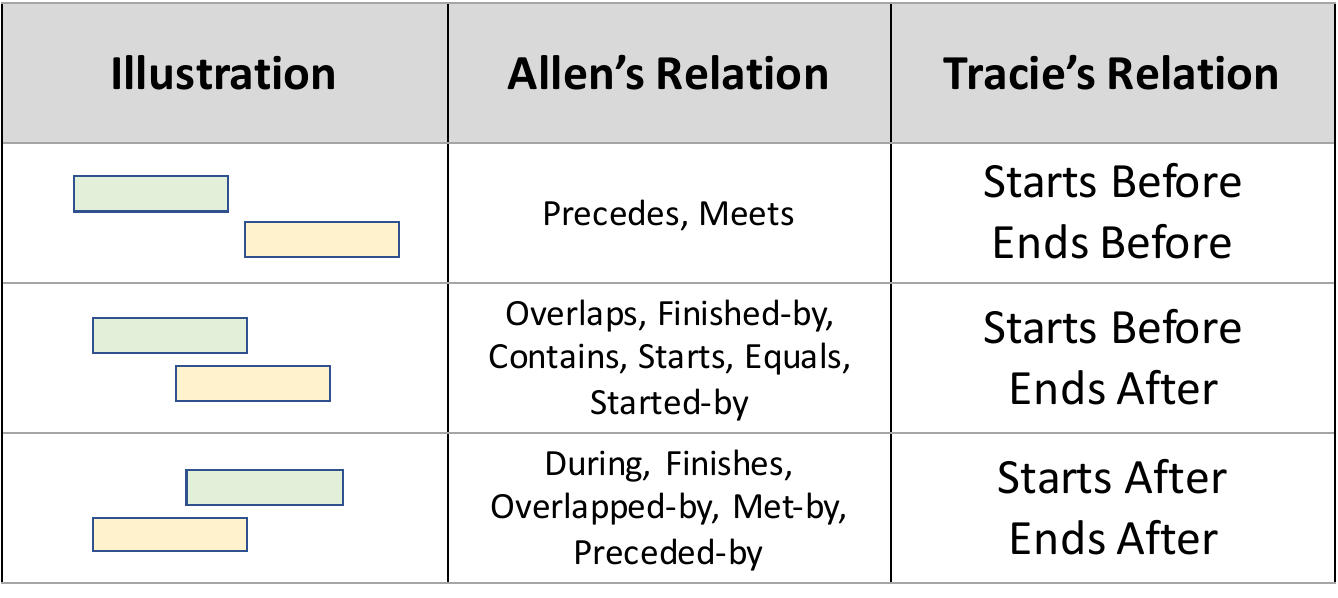}
    \caption{
        \datasetname{}'s label definition and its relation to Allen's interval algebra, with a graph illustration between an \textcolor{implicit-color}{implicit event} and an \textcolor{explicit-color}{explicit event}. 
    }
    \label{fig:allen-mapping}
\end{figure}

Since the start times of \textit{explicit-events} are more obvious to human annotators, we use them as reference points and compare the \textit{implicit-event}'s start or end time with them (depending on the \textit{comparator}), according to the label definitions shown in Fig.~{\ref{fig:allen-mapping}}.
In rare cases where two time points are the same (e.g., \textit{hit} and \textit{get hit} start at the same time in Fig.\ref{fig:timeline-example}), we use the causal relation to decide the order, so that \textit{hit} starts before \textit{get hit}.
Such instances are created through a multi-stage annotation process as detailed (in respective order) below. All steps are implemented with the CrowdAQ platform \cite{Qiang20Crowdaq} with qualification exams.


\paragraph{Implicit Event Generation} 

We randomly sample short stories from the ROCStories dataset \cite{mostafazadeh2016corpus}. For each story, one annotator writes 5 implicit event phrases that are not explicitly mentioned by the given story, but are inferable and relevant. The annotator additionally rewrites two explicit events closest to the implicit event's start and end time, respectively. With these two events, we can build two \datasetname{} instances (minus the \textit{temporal-relation}) per implicit event, which accounts for 10 instances in total per story.


\paragraph{Automatic Instance Generation} 
We use AllenNLP \cite{Gardner2018AllenNLPAD} to extract all verbs and relevant arguments with its semantic role labeling (SRL) model. With all the verbs and their arguments, we construct a pool of explicit events in the form of short phrases. For each implicit event, we randomly select two \{\textit{explicit-event}, \textit{comparator}\} pairs from the pool and build 10 additional instances (without \textit{temporal-relation}).


\paragraph{Label Collection}
For each of the 20 instances per story, we annotate the \textit{temporal-relation} with four different annotators. Annotators follow the label definition in \S\ref{subsec:tracie-overview} to produce four \textit{temporal-relation}s for each instance. We use the majority agreement as the final label and filter out unagreeable instances. Two authors additionally verify the instances with ambiguous verbs (e.g., ``have'') and corrected $5$\% of the end-time instances.

\subsection{Splits and Analysis}
\label{sec:splits}
We split the data under the independent and identically distributed (i.i.d.) assumption based on stories, with a 20/80 train/test ratio. We use a small training set, following \citet{zhou2019going}, as we believe temporal relations involve much commonsense knowledge. As we later show in \S\ref{sec:results}, it is infeasible to collect a large enough human-annotated training set to capture all the knowledge needed to tackle this problem completely, and a system must acquire knowledge from external resources. As a result, we use a small training set just to define the task, and at the same time, use an extensive testing set for more robust evaluation. 

The authors conduct a human upper-bound analysis on 100 randomly sampled instances, following the procedure in \citet{zhou-etal-2020-temporal}. There is a $94$\% agreement and a $98$\% resolved accuracy,\footnote{This is obtained after the authors discuss and resolve any disagreements before comparing with the annotated labels.} suggesting that \datasetname{} has a high annotation quality.


\section{Pattern-Based Pre-Training}
\label{sec:pattern-based-pretraining}
As argued in \S\ref{sec:splits}, we believe that it is more efficient to build a model that learns the prior knowledge needed for the task with distant signals and only subsequently learns the task definition through a small training set.
This section describes how we collect the distant signals related to events' start-time comparisons and pre-train a novel \emph{temporally-aware} transformer model called \modelpattern{}.
While \modelpattern{} will be used for fine-tuning directly on \datasetname{}, it will also form the basis of a more general temporal reasoning model called \modelsymbolic{} that we describe in \S\ref{sec:symtime}.

\subsection{Distant Supervision Collection}
\label{sec:starting-point-construction}
We describe the sources of distant supervision signals with the goal of understanding the relative order between two events' start times as well as the relative distance between them.

\begin{figure}[!ht]
    \centering
    \includegraphics[scale=0.58]{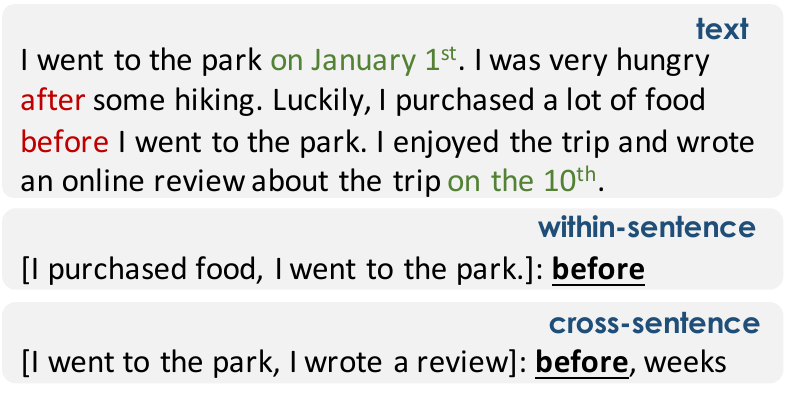}
    \caption{
        Extraction for start-time comparisons applied to an example paragraph. 
    }
    \label{fig:extraction-example}
\end{figure}

\paragraph{Within-Sentence Extraction}
We collect start time comparisons between pairs of events heuristically from free-text using ``before/after" keywords (following much prior work in temporal modeling and extraction \cite{Do2012JointIF}). We use AllenNLP's SRL model to process each input sentence and find verbs with a temporal argument that starts with either ``before'' or ``after'', and contains at least another verb. If there are multiple verbs in the temporal argument, we take the one with the largest number of tokens as arguments. We match the two extracted verbs with the relation indicated by the first word of either ``before'' or ``after''. 
As the example in Fig.~\ref{fig:extraction-example} shows, the extractor identifies that \textit{purchase food} is before \textit{go to park} as indicated by the ``before'' keyword mentioned in the text. 
We acquire 2.8 million instances from the May 2020 Wikipedia dump using this process.

\paragraph{Cross-Sentence Extraction}
\label{sec:cross-sentence-extraction}
The data collected from the within-sentence patterns does not reveal the relative distance between two start times. In addition, because writers often save trivial inferences for efficiency, certain event pairs rarely co-occur within a small textual window, making one event often implicit to the other one in these pairs.
To better collect such signals, we employ a cross-sentence extraction that finds direct temporal expressions of hours and dates. Because these temporal expressions (e.g., \keywordCode{2021-01-01}) are globally comparable, the compared events can be anywhere in a document. Therefore, this process collects more supervision signals about \textit{time-point comparisons} and their \textit{relative distance} on event pairs with trivial causal relations. We apply the SRL model and find all temporal arguments and their associated verbs. We find the exact temporal values by filling unmentioned elements of a temporal expression with the nearest previous mention (e.g., we add ``January'' to the expression of ``the 10th'' in Fig.~\ref{fig:extraction-example}.) These extractions have high precision, as the SRL model does well on identifying temporal arguments.

We then construct supervision instances under the assumption that the extracted temporal expressions describe the start times of the associated verbs (e.g., \textit{went} started on \textit{January $1^{\textrm{st}}$} in Fig.~\ref{fig:extraction-example}) . Each instance comprises an event pair, a temporal relation, and an estimation on the temporal difference between the two start times. Each event is a phrase constructed by taking all relevant arguments of the predicate verb in the SRL parses. We represent the differences between the two start times as one of seven coarse temporal units: 
\{$\leq$minutes, hours, days, weeks, months, years, $\geq$decades\}.
For example, we get \textit{go to park} is \textbf{weeks} before \textit{write review} as shown in Fig.~\ref{fig:extraction-example}. In addition to the event pairs, we randomly sample sentences within the paragraph to use as the context that better defines the events.
We collect 700k instances from this cross-sentence extraction process from Wikipedia.

\paragraph{Language Model (LM) Pre-Training Data}
We couple the specialized temporal pre-training data described above with additional paragraphs that are used to perform conventional language model pre-training using the original denoising task proposed in \citet{raffel2020exploring}. This is done to maintain part of the original language model's semantics and to avoid overfitting. We use the Gutenberg Dataset \cite{lahiri:2014:SRW} as the source and collect 1 million paragraphs for this purpose.


\paragraph{Data Format}
We then format the within / cross-sentence extraction data to consistent instances that have input sequences of \keywordCode{event:[EventA] starts [Relation][EventB].story:[Paragraph]} and output sequences of \keywordCode{answer:[Label][Distance]}. Here \keywordCode{[EventA]} represents the tokens that describe the first event; \keywordCode{[EventB]} represents the ones that describe the second event; and \keywordCode{[Paragraph]} represents the tokens of the context, which is non-empty only for cross-sentence extractions. \keywordCode{[Relation]} is either \keywordCode{before} or \keywordCode{after}, and \keywordCode{[Label]} is either \keywordCode{positive} or \keywordCode{negative}. When the label is positive, the relation will be the gold relation extracted from the text; when it is negative, the relation will be the inverse of the extracted relation. We randomly make 50\% of the instances negative. \keywordCode{[Distance]} is one of the 7 coarse temporal units represented with a set of blank tokens \keywordCode{[extra\_id\_N]}. We leave it to be blank for the within-sentence extractions so that the objective function will not include it in loss computations. The LM pre-training data follows the original format in \citet{raffel2020exploring}.

\subsection{Pattern-Based Temporal Model (\modelpattern{})}
\label{sec:pre-training-details}



We use a pre-trained sequence-to-sequence model as our base model and additionally pre-train this model using the data collected in \S\ref{sec:starting-point-construction} (for modeling details, see \S\ref{sec:baselines-and-systems}). We call the resulting model \emph{\modelpattern{}}. As a result of this additional pre-training step,  \modelpattern{} serves as new set of \emph{temporally-aware} model weights that can be used in place of existing pre-trained models and fine-tuned on \datasetname{}. As we describe next, we also use \modelpattern{} to build a modular temporal reasoning model called \modelsymbolic{} that attempts to go beyond a standard language modeling approach and improve start and end point prediction.



\section{Symbolic Temporal Reasoning Model (\modelsymbolic{})}
\label{sec:symtime}

To address the challenge of predicting event end times for which it is difficult to obtain high-quality direct or distant supervision, we introduce a new reasoning model called \emph{\modelsymbolic{}} in this section. This model makes end-time comparisons by symbolically combining start time distance and duration from separate predictions based on some of the components introduced in the previous section. Different from \citet{Leeuwenberg2018TemporalIE} and \citet{vashishtha2019fine}, our model does not rely on explicit annotations on timepoints, but only relative comparisons between them.


\subsection{Formulation}
\label{sec:symbolic-formulation}


\begin{figure}[t]
    \centering
    
    \centering 
    {\footnotesize
    \begin{tabular}{c  l}
        \hline 
        \textbf{comparator $l$} & \multicolumn{1}{c}{\textbf{relation  }\textbf{$r_{l}$$(\text{\hlgray{e$_{1}$}},\text{\hlgray{e$_{2}$}})$=}} \\ \hline 
        \hl{\textbf{ends}} & {$
\begin{cases}
\hlcyan{\textbf{before}} & \text{if }\mathbf{end}_{1} < \mathbf{start}_{2} \\
\hlcyan{\textbf{after}} &\emph{otherwise}
\end{cases}
$}
 \\ 
        \hl{\textbf{starts}} & {$
\begin{cases}
\hlcyan{\textbf{before}} & \text{if }\mathbf{start}_{1} < \mathbf{start}_{2} \\
\hlcyan{\textbf{after}} &\emph{otherwise}
\end{cases}
$}
        \\ 
    \end{tabular}}
    \caption{Decomposition of the relation functions that solve \datasetname{} instances (equal timepoints ignored). 
    }
    \label{fig:sym_func}
\end{figure}

As described in \S\ref{subsec:tracie-overview}, hypotheses in \datasetname{} make pair-wise comparisons between two events $e_{1}$ and $e_{2}$ using a \emph{comparator} $l$ from $\{\texttt{starts},\texttt{ends}\}$ and a \emph{query-relation} $r$ from $\{\texttt{before}, \texttt{after}\}$ based on a provided story context. We associate each $e_j$ with a latent start time $\mathbf{start}_{j}$ and an end time $\mathbf{end}_{j}$, as well as, for convenience, a duration $\mathbf{duration}_{j} = \mathbf{end}_{j} - \mathbf{start}_{j}$. Under this formulation, a symbolic approach to solving \datasetname{} involves computing the \emph{relation functions} $r_{l}$ shown in Figure~\ref{fig:sym_func}. For example, given exact numeric values $\mathbf{end}_{1}$ and $\mathbf{start}_{2}$, as one would assume in a classical interval-based approach to temporal reasoning \cite{allen1983maintaining}\footnote{In the Allen algebra, the values $\mathbf{end}_{x}$ and $\mathbf{start}_{y}$ correspond to the right and left end points $x^{+}, y^{-}$ in the intervals $(x^{-},x^{+}), (y^{-},y^{+})$. Likewise, our $\mathbf{duration}_{x}$ corresponds to the value $(x^{+} - x^{-})$.}, determining if the first event \emph{ends before} the second involves simply computing whether $\mathbf{end}_{1}$ is less than $\mathbf{start}_{2}$.


Given that the exact values of start and end times are latent, we use the intervals to do the same comparisons, as they are more context-invariant. For example, we do not need the exact date to know that \textit{lunch} starts before \textit{dinner} in the same day, because there is a typical distribution of the relative distance between the two start times. Based on this idea, we build a neural-symbolic model that learns approximations of these simple functions in Fig.~\ref{fig:sym_func} in a differentiable way. Specifically, we use individual neural modules that make predictions about event intervals via distance and duration functions $\mathrm{dist}(e_{i},e_{j})$ and $\mathrm{dur}(e_{j})$, respectively.


To understand this decomposition, we define the distance and duration functions computed by these two modules as $\mathrm{dist}(e_i, e_j) = \mathbf{start}_{i} - \mathbf{start}_{j}$ and $\mathrm{dur}(e_{j}) = \mathbf{duration}_{j}$. By exploiting the rule that an end point $\mathbf{end}_{j}$ can be computed as $\mathbf{end}_{j} = \mathbf{start}_{j} + \mathbf{duration}_{j}$, we can, for example, decompose the relation $r_{ends}(e_{1},e_{2}) = \textbf{before}$ (i.e., \emph{$e_{1}$ ends before $e_{2}$}) in terms of our two modules as follows via simple algebraic manipulation:
\begin{align*}
    &r_{ends}(e_{1},e_{2}) = \textbf{before} \\
    &\Leftrightarrow\ \mathbf{end_1} < \mathbf{start_{2}}  \\ 
    &\Leftrightarrow\ \mathbf{start_{1}} + \mathbf{duration_{1}} < \mathbf{start_{2}} \\
    &\Leftrightarrow\ \big( \mathbf{start_{1}} - \mathbf{start_{2}}\big) + \mathbf{duration_{1}} < 0 \\
    &\Leftrightarrow\ \mathrm{dist}(e_1, e_2)+ \mathrm{dur}(e_1) < 0
\end{align*}
Hence, we have reduced the computation of the relation \emph{ends before} to a \emph{symbolic computation} over two numeric intervals.
Conversely, we have $r_{ends}(e_{1},e_{2}) = \textbf{after}  \Leftrightarrow\ \mathrm{dist}(e_1, e_2)+ \mathrm{dur}(e_1) > 0$,\footnote{We note that one drawback of this inference rule is that it does not predict causal relations and, therefore, cannot handle instances where $\mathbf{end_1}=\mathbf{start_2}$ as our label definitions describe in \S\ref{subsec:tracie-overview}. We leave this problem for future research.}
For the \textbf{starts} comparator, we have $r_{starts}(e_{1},e_{2}) = \textbf{before}  \Leftrightarrow\ \mathrm{dist}(e_1, e_2) < 0$ and vice versa for the \textbf{after} relation.

\begin{figure}[t]
    \centering
    \includegraphics[scale=0.5]{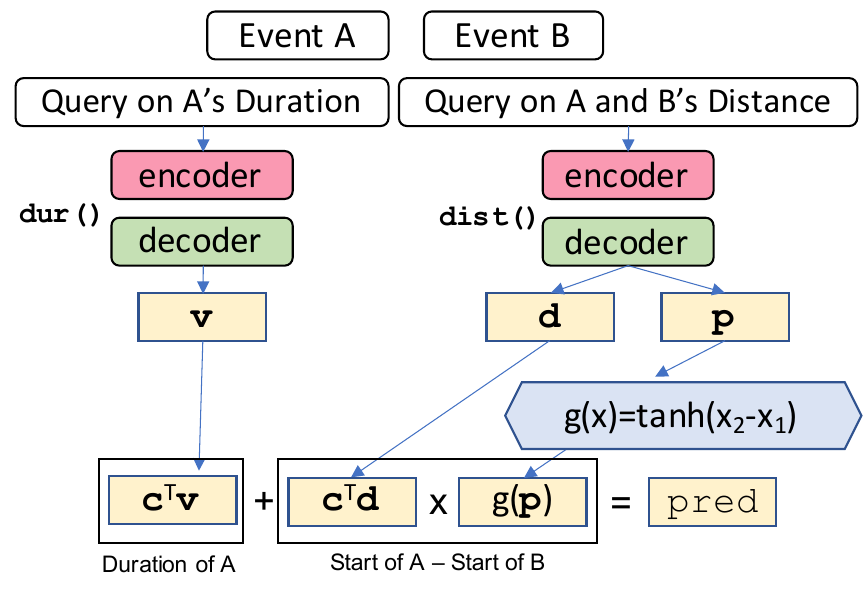}
    \caption{
        A schematic overview of \modelsymbolic{} 
        to compare event $A$'s end time with event $B$'s start time
        via modular predictions about $A$'s duration and distance from $B$ and their symbolic combination (bottom).  
    }
    \vspace{-0.2cm}
    \label{fig:model-overview}
\end{figure}

In what follows, we describe how we approximate the values of the two functions via individual neural modules (see illustration in Fig.~\ref{fig:model-overview}).

\subsection{Duration Estimation}
\label{sec:duration-estimation}
To obtain a model to estimate $\mathrm{dur}(\cdot)$, we pre-train a sequence-to-sequence model with the duration data from \citet{zhou-etal-2020-temporal}, which is similarly collected from pattern-based extraction. The data contains over 1 million events with their corresponding duration values. We map each instance to an input sequence \keywordCode{event:[Event]story:[Story]} and a corresponding output sequence \keywordCode{answer:[Value]}, where \keywordCode{[Event]} represents the tokens of an event with the trigger verb marked by a special token to its left, \keywordCode{[Story]} represents down-sampled tokens from the context, and \keywordCode{[Value]} is one of the 7 unit labels as described in \S\ref{sec:cross-sentence-extraction} (i.e., \{ $\leq$\text{minutes}, \text{hours}, \text{days}, \text{weeks}, \text{months}, \text{years}, $\geq$\text{decades} \}). 


\subsection{Computation and Learning}
\label{sec:symbolic-computation}
We use the output from \modelpattern{} to approximate the function $\mathrm{dist}(\cdot)$.
Following the sequence formulation of \modelpattern{} in \S\ref{sec:pattern-based-pretraining}, we replace \keywordCode{[EventA]} with the textual description of $e_1$, \keywordCode{[EventB]} with the textual description of $e_2$, and \keywordCode{[Paragraph]} with the context (premise), and fix \keywordCode{[Relation]} to be \textit{before}. 
By taking the values of the vocabulary indices corresponding to ``positive'' and ``negative'' from the logits of \keywordCode{[Label]}
and applying a $\mathrm{softmax}$ operation, we get $P_{\textrm{before}}$ and $P_{\textrm{after}}$. These are the probability of $e_1$ starting before and after $e_2$, respectively, and are used to define the vector $\vb{p}=[P_{\textrm{before}}, P_{\textrm{after}}]$.
Similarly, we apply $\mathrm{softmax}$ to the logits of \keywordCode{[Distance]} over the 7 words representing the temporal units 
to obtain 7 values that approximate the probabilities of the 
distance between two events' start times being closest to each temporal unit. We place the 7 values in temporal units' increasing order in vector $\vb{d}$.
To represent $|\mathbf{start_1}-\mathbf{start_2}|$ with a single value, we dot product the probabilities with an incremental constant vector $\vb{c}=[0,1,2,3,4,5,6]$.
To get the direction, we apply the $\mathrm{tanh}$ function to the difference between the probabilities in $\vb{p}$.\footnote{To ensure that $\mathrm{tanh}$ returns a value close to 1 or -1, we multiply the distance by a big number denoted as $\textrm{INT}_{max}$.} As a result, we have:

\begin{equation} \label{eq-start-module}
\begin{split}
\mathrm{dist}(\cdot) &= \mathbf{start_1} - \mathbf{start_2} \\
 &= \vb{c}^T\vb{d}*\mathrm{tanh}(\textrm{INT}_{max}*(\mathbf{p}_2-\mathbf{p}_1)) \\
\end{split}
\end{equation}

We use the pre-trained model in \S\ref{sec:duration-estimation} to approximate the function $\mathrm{dur}(\cdot)$. Because the model is pre-trained with markers to the left of trigger verbs, we run a part-of-speech tagger on input phrases and add a marker to the left of the first verb.
We apply $\mathrm{softmax}$ to the logit values of \keywordCode{[Value]} over the 7 temporal unit words 
and get, as above, 7 values representing the probabilities of the input event's duration being closest to each unit. We form $\vb{v}$ by placing these values at the temporal unit's increasing order. With the same constant vector, we have:
\begin{equation}
\mathrm{dur}(\cdot) = \mathbf{duration_1} = \vb{c}^T\vb{v}
\end{equation}

For hypotheses with comparator $\texttt{starts}$, we use \modelpattern{} and its sequence-to-sequence objective to learn (i.e., we take the input hypothesis and context as is and use \keywordCode{[Label]} directly as the prediction). For hypotheses where the comparator is $\texttt{ends}$, we use the inference process in \S\ref{sec:symbolic-formulation} and the computation process described above to construct $logits=[\mathbf{pred}, -\mathbf{pred}], \mathbf{pred}=\mathrm{dist}(e_1, e_2)+ \mathrm{dur}(e_1)$ as detailed in Fig.~\ref{fig:model-overview}.
We find the \textit{gold-temporal-relation} in each training instance and compute a two-class cross-entropy loss with $logits$. The \modelpattern{} that predicts $\texttt{starts}$ hypotheses shares weights with the one used in computing $logits$. The final model \modelsymbolic{} can also be used to predict \datasetname{} instances without any task-specific supervision as the two functions are initialized with distant supervision.

\section{Experiments}
\label{sec:experiments}

In this section, we detail our experimental setup (\S\ref{sec:baselines-and-systems}-\ref{sec:metrics}) and report our main results (\S\ref{sec:results}-\ref{sec:ablation-studies}).\footnote{We release the systems for reproduction at \url{http://cogcomp.org/page/publication_view/937}}




\subsection{Baselines and Systems}
\label{sec:baselines-and-systems}
We use T5-Large implemented by \citet{wolf2019transformers} as our base sequence-to-sequence model for both \modelpattern{} and the duration model in \S\ref{sec:duration-estimation} as it provides for faster iterations. 
We use early stopping, batch size of 32 and other default parameters. \modelpattern{} converges after 45k steps ($\sim$1.4M instances) and the duration model converges after 80k steps ($\sim$2.6M instances). We use these pre-trained weights in \modelsymbolic{} as well as \modelzeroshot{} which uses no \datasetname{} supervision.

We compare with our proposed models with a host of baselines based on the same pre-trained language model, including \textbf{BaseLM}: T5-Large, and \textbf{BaseLM-MATRES}: T5-Large fine-tuned on 20k MATRES training data. We also compare with other architectures/models, including \textbf{BiLSTM} as used in \citet{williams2017broad}, \textbf{Roberta-Large} \cite{Liu2019RoBERTaAR} and \textbf{T5-3B}. All models and baselines follow a standard TE setup and default parameters. We report a 3-run average and each model is run until convergence.



\subsection{Metrics and Settings}
\label{sec:metrics}
We measure system performance on \datasetname{} separately for start-time hypotheses and end-time hypotheses. We also employ a story-wide exact match metric, which is the percentage of stories with all its related hypotheses answered correctly.


In addition to \datasetname{}'s standard i.i.d.\ split, we propose a pruned version of the training set with balanced prior distributions. For example, in the i.i.d.\ training set, 70\% of the examples with the comparator \texttt{ends} and relation \textit{after} are positive. We randomly remove instances from the majority classes to produce a uniform-prior training set such that a model can no longer rely on such prior distributions. We believe this setting better evaluates a system's true understanding of the task.


\subsection{Main Results}
\label{sec:results}
\begin{table}[t]
\small
\centering
    \begin{tabular}{l|ccc|c}
        \toprule 
        System                                     &\ Start\ \ &\ End\ \ &\ All\ &\ Story\ \\ 
        \cmidrule(lr){1-1}                              \cmidrule(lr){2-2} \cmidrule(lr){3-3} \cmidrule(lr){4-4} \cmidrule(lr){5-5}
        Majority & 57.3 & 69.8 & 64.1 & 18.1 \\
        BiLSTM & 53.7 & 63.5 & 59.1 & 10.9\\ \cdashline{1-5}
        Roberta-Large & 78.5 & 78.3 & 78.4 & 26.1\\
        T5-3B & 79.4 & 77.4 & 78.3 & 26.9 \\
        \midrule
        BaseLM (T5-large) & 75.5 & 75.4 & 75.4 & 22.6 \\
        BaseLM-MATRES & 76.7 & 76.3 & 76.5 & 25.3 \\ \cdashline{1-5}
        \modelpattern{} (ours) & 81.4 & 77.5 & 79.3 & 31.0 \\
        \modelsymbolic{} (ours) & \textbf{82.1} & \textbf{79.4} & \textbf{80.6} & \textbf{32.0} \\ \cmidrule(lr){1-5}
        \modelzeroshot{} & 77.0 & 73.1 & 74.9 & 21.6 \\
        \bottomrule 
    \end{tabular}
\caption{
Performance on \datasetname{}, best numbers in \textbf{bold}. BaseLM is T5-large; Story is the percentage of story-wide exact match; Majority is based on the comparator and temporal-relation distribution; Zeroshot uses no \datasetname{} instance as supervision.}
\label{tab:performance-iid-main}
\end{table}

\begin{table}[t]
\small
\centering
    \begin{tabular}{l|ccc|c}
        \toprule 
        System                                     &\ Start\ \ &\ End\ \ &\ All\ &\ $\Delta$All\ \\ 
        \cmidrule(lr){1-1}                              \cmidrule(lr){2-2} \cmidrule(lr){3-3} \cmidrule(lr){4-4} \cmidrule(lr){5-5}
        Random & 50.0 & 50.0 & 50.0 & -14.1 \\
        BiLSTM & 50.5 & 51.2 & 50.9 &  -8.2 \\ \cdashline{1-5}
        Roberta-Large & 75.1 & 68.1 & 71.3 & -7.1 \\ 
        T5-3B & 72.8 & 68.6 & 70.5 & -7.8 \\
        \midrule
        BaseLM (T5-large) & 68.1 & 67.8 & 67.9 & -7.5 \\
        BaseLM-MATRES & 76.3 & 69.9 & 72.8 & -3.7 \\ \cdashline{1-5}
        \modelpattern{} (ours) & 80.6 & 73.2 & 76.6 & -2.7\\
        \modelsymbolic{} (ours) & \textbf{81.2} & \textbf{77.0} & \textbf{78.9} & -1.7 \\ \cmidrule(lr){1-5}
        \modelzeroshot{} & 77.0 & 73.1 & 74.9 & \textbf{0.0} \\
        \bottomrule 
    \end{tabular}
\caption{
Performance on \datasetname{} uniform-prior training setting. $\Delta$All compares the difference with Table~\ref{tab:performance-iid-main}; Majority is equivalent to random guessing.}
\label{tab:performance-iid-balanced}
\end{table}

Table~\ref{tab:performance-iid-main} shows system performance on \datasetname{}'s i.i.d.\ setting.
We observe that \modelpattern{} improves on all metrics over the base language model, with 6\% on start-time comparisons and 8\% on story-wide exact match. It also outperforms BaseLM-MATRES, suggesting that distant supervision is more efficient than extensive human annotation. 

With a symbolic end-time inference, \modelsymbolic{} further improves on all metrics, with 7\%, 4\%, and 9\% gains over the base language model on start time, end time and story-wide exact match, respectively. \modelsymbolic{} can further improve the performance on start-time hypotheses over \modelpattern{} even though they use the same model to predict start-time queries. This is because \modelpattern{} is not designed to understand end time from pre-training, and fine-tuning on such data hurts its representation in general. This illustrates the benefits of models using explicit and sensible reasoning processes.

Table~\ref{tab:performance-iid-balanced} compares systems in the uniform-prior training setting. Compared to the setting in Table~\ref{tab:performance-iid-main}, a system cannot exploit prior knowledge about the label distribution when making predictions. Given this, we see that all baselines produce a much lower performance, e.g., the BiLSTM, which is a model that lacks much of the pre-requisite knowledge for reasoning, suddenly performs near random chance. Compared to the baseline models, \modelpattern{} only drops $2.7$\%, suggesting that it is more invariant to evaluation settings and better understands temporal common sense.
\modelsymbolic{} has the smallest drop among all models (1.7\%) because of its explicit reasoning process on end-time hypotheses. \modelzeroshot{} does not use any \datasetname{} training examples, so it has the same performance in the uniform-prior setting which outperforms all supervised baselines including T5-3B.

\subsection{Extrinsic Evaluation}
\label{sec:extrinsic-evaluation}

\begin{table}[t]
\small
\centering
    \begin{tabular}{l|c|c|c|c}
        \toprule 
        System                                     & OT-NS & OT & OT-MS & PT \\ 
        \cmidrule(lr){1-1}                              \cmidrule(lr){2-2} \cmidrule(lr){3-3} \cmidrule(lr){4-4} \cmidrule(lr){5-5}
        \citet{WCZR20} & 85.9 & - & - & - \\
        BaseLM & 86.0 & 87.5 & 77.4 & 69.0  \\
        \modelsymbolic{} & 87.3 & 89.6 & 86.1 & 75.1 \\
        \bottomrule 
    \end{tabular}
\caption{Performance on MATRES. \citet{WCZR20} is not strictly comparable with the rest.}
\vspace{-1em}
\label{tab:performance-matres}
\end{table}
To show that our model is not limited to the \datasetname{} dataset and is general in temporal relation reasoning, we also evaluate on MATRES \cite{NingWuRo18}, a temporal relation dataset focused on comparing explicit events' start times. We train and evaluate only the instances with a label of either ``before'' or ``after'', which accounts for about 80\% of all instances.
We compare the performance of \modelsymbolic{}\footnote{This is virtually the same as using \modelpattern{} as MATRES does not evaluate duration nor end times.} with BaseLM. 
We report four results - \textbf{OT-NS (original test, no story)}: train and test with only the sentences containing the trigger verbs; \textbf{OT}: train and test with the entire document (down-sampled to be below the maximum sequence length) as an auxiliary input; \textbf{OT-MS (original test, minimal supervision)}: train with 1.2k (6\%) training instances; \textbf{PT (perturbed test)}: train with the complete training set and test on a perturbed test set from \citet{Gardner2020EvaluatingNM}. In OT-NS, we also report a SOTA system from \citet{WCZR20} under the same two-label\footnote{\citet{WCZR20} is trained with two additional labels. 
We constraint the output space to only ``before'' and ``after'' using argmax, but this process makes it not directly comparable.} setting.

Table~\ref{tab:performance-matres} shows the performance of our model and the baselines. We see that our model is consistently better than BaseLM, and at the same time, comparable to \citet{WCZR20}. Our model benefits more from input contexts, and only drops 4\% in the OT-MS setting with minimal supervision (from 89.6 to 86.1), comparing to the 10\% drop from T5-Large. This shows the effectiveness of our distant signals in \S\ref{sec:starting-point-construction}, which are also designed to encourage contextual understandings.

\subsection{Ablation Studies and Analysis}
\label{sec:ablation-studies}

\noindent To better understand the improvements from our models, we conduct several ablation studies. 

\begin{table}[t]
\small
\centering
    \begin{tabular}{c|c|c|c|c}
        \toprule 
        Sys.                                    & BaseLM & \modelpattern{} & \modelsymbolic{} & Human\\ 
        \cmidrule(lr){1-1} \cmidrule(lr){2-2} \cmidrule(lr){3-3} \cmidrule(lr){4-4} \cmidrule(lr){5-5}
        Acc. & 52.6 & 72.2 & 75.3 & 82.5\\
        \bottomrule 
    \end{tabular}
\caption{Performance on \textit{no-story} \datasetname{} under the uniform-prior training setting.}
\vspace{-1em}
\label{tab:performance-ablation-hyponly}
\end{table}

Table~\ref{tab:performance-ablation-hyponly} shows the results on \datasetname{} where the story is not provided as part of the inputs to systems (a \emph{no-story} setting). While such a setting bares some resemblance to the \emph{partial-input} baselines often employed in TE \cite{poliak2018hypothesis}, in our setting, it is often possible to predict temporal relations in the absence of stories because of strong commonsense priors.
Indeed, we estimate that $65$\% of the instances can be correctly predicted from the hypotheses alone, based on expert analysis in \S~\ref{sec:splits}. This suggests a $82.5$\% human upper-bound\footnote{We assume that the remaining $35$\% non-predictable instances are decided by random guessing.} in this \textit{no-story} setting. Hence, such a setting partly evaluates a model's ability to incorporate commonsense priors when making decisions.

We see that BaseLM is close to random chance, whereas \modelpattern{} and \modelsymbolic{} improve 20\% and 22\% respectively. This suggests that our models better understand temporal common sense through the distant supervision on both start times and duration. On the other hand, we observe much smaller drops in our model's performances in this \textit{no-story} setting. This suggests that our models do not improve as much on the 35\% instances that require multi-hop timeline constructions over more than two events, motivating future work.

\begin{table}[t]
\small
\centering
    \begin{tabular}{c|c|c|c}
        \toprule 
        Sys.                                    & \modelpattern{} & cross-sentence & within-sentence \\ 
        \cmidrule(lr){1-1} \cmidrule(lr){2-2} \cmidrule(lr){3-3} \cmidrule(lr){4-4}
        Acc. & 80.6 & 79.9 & 63.7\\
        \bottomrule 
    \end{tabular}
\caption{Comparison of pre-training data sources on \datasetname{}'s start time prediction accuracy, under the uniform-prior training setting.}
\vspace{-1em}
\label{tab:performance-ablation-source}
\end{table}

Table~\ref{tab:performance-ablation-source} compares the two pre-training sources described in \S\ref{sec:starting-point-construction} by individually pre-training two models with only within-sentence or cross-sentence extracted data. We see that the cross-sentence extraction brings the most performance gain on \datasetname{}'s start-time binary metric under the uniform-prior training setting. This suggests that the global extraction rule is able to introduce new knowledge that is not seen in localized language model pre-training. Combining the within-sentence data further improves the performance. 

Through analysis on the interval predictions made by \modelsymbolic{}, we notice a tendency for the model to predict ``after'' for end-time instances, possibly due to overly-estimated durations: a byproduct of natural biases in text. Given the weak signal used to learn such intervals and these potential biases, this is not altogether surprising. We leave the task of learning more robust and faithful interval representations for future work. 


\section{Conclusion}
We introduce a challenging dataset \datasetname{}, to evaluate systems' temporal understanding of implicit events. We propose a distant supervision process that improves language models' understanding of start times of both explicit and implicit events. We further combine this process with a distantly supervised model that estimates events' duration to compare event end times, under the explicit rule that end times are start times plus durations. We show that our model improves over \datasetname{} and MATRES, suggesting the effectiveness of high-precision pre-training and symbolic temporal reasoning. Despite these advances, \datasetname{} continues to be a challenging task for future work on general temporal reasoning.


\section*{Acknowledgments}
This research is based upon work supported in part by the office of the Director of National Intelligence (ODNI), Intelligence Advanced Research Projects Activity (IARPA), via IARPA Contract No. 2019-19051600006 under the BETTER Program, and by Contract FA8750-19-2-1004 with the US Defense Advanced Research Projects Agency (DARPA). The views expressed are those of the authors and do not reflect the official policy or position of the Department of Defense or the U.S. Government.

\bibliographystyle{acl_natbib}
\bibliography{tracie}

\end{document}